\documentclass{article}

\usepackage{graphicx}
\usepackage{subfigure}
\usepackage{booktabs} 
\usepackage{enumitem}
\usepackage{pgfplots}
\usepackage{listings}
\usepackage{color}
\usepackage{xcolor} 

\definecolor{NL4OPTcolor}{HTML}{6C8EBF}
\definecolor{ComplexORcolor}{HTML}{B85450}
\definecolor{NLP4LPcolor}{HTML}{82B366}

\definecolor{NL4OPTcolorLight}{HTML}{DAE8FC}
\definecolor{ComplexORcolorLight}{HTML}{F8CECC}
\definecolor{NLP4LPcolorLight}{HTML}{D5E8D4}

\pgfplotsset{compat=1.15}

\usepackage{algorithmicx}
\usepackage{algorithm}
\usepackage{algpseudocode}
\algtext*{EndIf}
\algtext*{EndFor}

\newcommand{\pluseq}{\mathrel{+}=}

\usepackage{hyperref}

\usepackage[accepted]{icml2024}

\usepackage{amsmath}
\usepackage{amssymb}
\usepackage{mathtools}
\usepackage{amsthm}

\usepackage[capitalize,noabbrev]{cleveref}

\theoremstyle{plain}

\theoremstyle{definition}

\theoremstyle{remark}

\usepackage[textsize=tiny]{todonotes}

\newcommand{\revised}[1]{#1}

\icmltitlerunning{Large Language Model for Optimization Modeling}

\setlist{itemsep=0.2ex, topsep=1ex, partopsep=0ex, parsep=0.5ex, leftmargin=2.3ex}

\begin{document}

\twocolumn[
\icmltitle{OptiMUS: Scalable Optimization Modeling \\ with (MI)LP Solvers and Large Language Models}



\icmlsetsymbol{equal}{*}

\begin{icmlauthorlist}
\icmlauthor{Ali AhmadiTeshnizi}{mse}
\icmlauthor{Wenzhi Gao}{icme}
\icmlauthor{Madeleine Udell}{mse,icme}
\end{icmlauthorlist}
\vspace{1em}

\definecolor{NL4OPTcolor}{HTML}{6C8EBF}
\definecolor{ComplexORcolor}{HTML}{B85450}
\definecolor{NLP4LPcolor}{HTML}{82B366}

\icmlaffiliation{mse}{Department of Management Science and Engineering, Stanford University, CA, USA}
\icmlaffiliation{icme}{Institute for Computational and Mathematical Engineering, Stanford University, CA, USA}

\icmlcorrespondingauthor{Ali AhmadiTeshnizi}{teshnizi@stanford.edu}

\icmlkeywords{Machine Learning, ICML}

]



\printAffiliationsAndNotice{}  

\begin{abstract}

Optimization problems are pervasive in sectors from manufacturing and distribution to healthcare.
However, most such problems are still solved heuristically by hand rather than optimally by state-of-the-art solvers because the expertise required to formulate and solve these problems limits the widespread adoption of optimization tools and techniques.
This paper introduces OptiMUS, a Large Language Model (LLM)-based agent designed to formulate and solve
(mixed integer) linear programming problems from their natural language descriptions. 
OptiMUS can develop mathematical models, write and debug solver code, evaluate the generated solutions, and improve its model and code based on these evaluations. 
OptiMUS utilizes a modular structure to process problems, allowing it to handle problems with long descriptions and complex data without long prompts. 
Experiments demonstrate that OptiMUS outperforms existing state-of-the-art methods on easy datasets by more than $20\%$ 
and on hard datasets (including a new dataset, NLP4LP, released with this paper that features long and complex problems) 
by more than $30\%$. 

\end{abstract}

\begin{figure*}[t]
    \centering
    \includegraphics[width=0.999\textwidth]{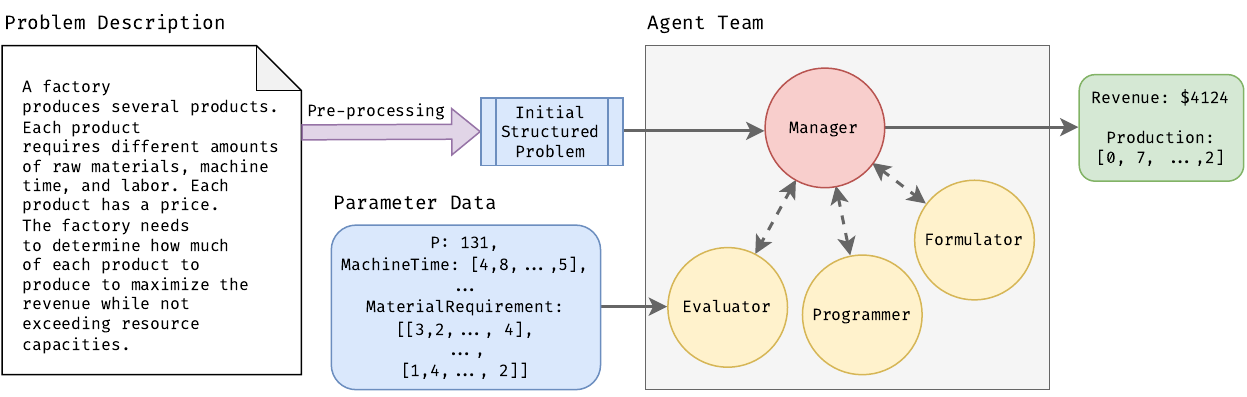}
    \caption{
     OptiMUS uses a structured sequence of LLM agents to effectively model and solve optimization problems. First, the natural language representation of the problem is preprocessed into a structured problem. Then, a team of agents iteratively augments the structured problem with a connection graph, mathematical formulations of each clause, and code for each clause. The agents continue work until the problem is solved. (Dashed lines represent communications that can occur multiple times.)}
    \label{fig:agent-team}
\end{figure*}

\section{Introduction}
\label{intro}
Optimization problems are common in many fields such as operations, economics, engineering, and computer science. 
Important applications of optimization include reducing energy costs in smart grids, improving supply chains, and increasing profits in algorithmic trading \citep{OptModelingApp, antoniou2007practicalOptimization}. 
Major advances in optimization algorithms over the last several decades have led to reliable and efficient optimization
methods for a wide variety of structured optimization problems, including linear programming (LP) and mixed-integer linear programming (MILP), among many others. 
Unfortunately, optimization modeling, transforming a business problem into a mathematical optimization problem, still requires expert knowledge. According to a recent survey, 81\% of Gurobi's commercial solver users have advanced degrees, with 49 \% of them holding a degree in operations research \citep{gurobiReport}.
This expertise gap prevents many organizations from using optimization, even when it could significantly improve their operations. Examples include inventory management in supermarkets, patient operations in hospitals, transportation policies in small municipalities, energy management in local solar farms, and operations in small businesses or NGOs \citep{saghafian2015operationsHospital, aastrup2010fortyRetail, yao2020optimizationTransport, shakoor2016wakeFarm}. 
Automating optimization modeling would allow sectors that cannot afford to have access to optimization experts to improve efficiency using optimization techniques.  
Large language models (LLMs) offer a promising way to make optimization more accessible. 
LLMs have demonstrated the ability to understand, generate, and interpret natural language for many tasks.  In the optimization domain, LLMs can make it easier to formulate problems and obtain solutions, making expert-level optimization more accessible \cite{ramamonjison2023nl4opt}
However, several challenges remain before LLMs can reliably model real-life optimization problems:


\begin{itemize}[leftmargin=10pt]

    \item \textbf{Ambiguous Terms:} An optimization problem can be described in many ways. For example, a user might use different terms (e.g. vehicle vs. car vs. truck vs. carrier), notations (e.g. \textit{price} and \textit{capacity} vs. $p$ and $c$ vs. $x$ and $y$), or omit common-sense assumptions (e.g. capacity of a vehicle is non-negative, number of employees is an integer, etc.). 
    Moreover, defining the right variables can be a challenge. 
    For instance, information flow through a network requires a different set of variables than physical goods, as the quantity of information need not be conserved. 

    \item \textbf{Long Problem Descriptions:} LLMs have a limited context size. However, real-world problems can be long and complex: for example, the energy system problem in \cite{Challenge3} has a 60-page documentation.
    Even for long-context models, performance decreases substantially as the input context grows \citep{liu2023lostinthemiddle}. Consequently, LLMs tend to make more mistakes as the length of the problem description increases and perform poorly on complex problems.
    
    \item \textbf{Large Problem Data: } The specification of an optimization problem often involves large amounts of data, such as customer attributes or sales of goods. Previous approaches to optimization modeling using LLMs, which pass numerical data to the LLM directly, are thus restricted to the simplest of toy problems. 

    \item \textbf{Unreliable Outputs:} The solutions provided by LLMs are not always reliable. The generated code may be incorrect or even not executable. It is especially challenging to verify the solution when the code runs, but the output is incorrect. For instance, if the code runs and claims that the problem is unbounded, perhaps a constraint has been accidentally omitted from the formulation. 
\end{itemize}


\paragraph{Contributions.}
This paper develops a novel perspective on optimization modeling that addresses each of these limitations and makes the following contributions:

\begin{itemize}[leftmargin=10pt]
    \item Existing datasets for natural language optimization modeling are too easy to capture the challenge of long problem descriptions and large problem data. 
    This work introduces NLP4LP, an open source dataset of 67 complex optimization problems. \cref{table:datasets} compares NLP4LP to existing datasets and \cref{section:dataset} describes NLP4LP.

    \item We develop a modular, LLM-based agent to model and solve optimization problems, which we call OptiMUS. OptiMUS beats the previous state-of-the-art methods on existing datasets by over 20\%
    and on our more challenging dataset by 30\%. 
    OptiMUS employs a novel connection graph that allows it to process each constraint and objective independently. 
    Using this connection graph, and separating data from the problem description, 
    OptiMUS can solve problems with long descriptions and large data files without excessively long prompts.
    
    
    


    
        
\end{itemize}

\paragraph{Structue of the Paper}
This paper is organized as follows: \cref{background} discusses the background and related work; \cref{methodology} describes the details of our LLM-based optimization agent; \cref{experiments} discusses the datasets and presents the experiments and analysis; \cref{conclusion} concludes the paper with future directions and implications. The appendix includes prompts, details on the experiments' setup, and further analysis.

\section{Background and Related Work}
\label{background}

Optimization problems are mathematically defined by an objective function and a set of constraints. 
For example, an MILP can be written mathematically as
\begin{align}
  \underset{\{ x_j \}}{\text{minimize}}\quad  & \sum_{j = 1}^n c_j x_j
  \nonumber\\
  \text{subject to}\quad & \sum_{j = 1}^n a_{i j} x_j {~(\leq,=,\geq)~ b_i} , i = 1, \ldots m
  \nonumber\\
  & l_j \leq x_j \leq u_j, j = 1, \ldots, n \nonumber\\
  & x_j \in \mathbb{Z}, j \in \mathcal{I} \nonumber
\end{align}
An optimization workflow consists of \textbf{1)} formulating an optimization problem in mathematical form
by identifying its objective and constraints, and then 
\textbf{2)} solving the realization of problem from real data, generally using code that calls an optimization solver.



\begin{table*}[t]
\caption{A comparison on different aspects of complexity for various datasets. The unit for description length is characters}
\label{table:datasets}
\begin{center}
\begin{small}

\begin{tabular}{lccc}
\toprule
Dataset & Description Length & Instances (\#MILP)  & Multi-dimensional Parameters \\
\midrule
NL4Opt     & 518.0 $\pm$ 110.7 & 1101 (0) & $\times$ \\
ComplexOR & 497.1 $\pm$ 247.5 & 37 (12) & $\checkmark$\\
NLP4LP (Ours) & 908.9 $\pm$ 504.6 & 67 (13) & $\checkmark$\\
\bottomrule
\end{tabular}

\end{small}
\end{center}
\end{table*}

\paragraph{Progress in LLMs.} Recent progress in Natural Language Processing (NLP) has led to the development of large language models (LLMs) useful for tasks such as answering questions, summarizing text, translating languages, and coding.

\citep{openai2023gpt4, touvron2023llama, chowdhery2022palm, wei2023chainofthought,gao2023pal,borgeaud2022improving}. Connections to other software tools extend the reach and accuracy of LLMs, as demonstrated by plug-ins for code writing and execution \citep{paranjape2023art, wei2023chainofthought}.
\citep{yang2023LLMsAsOptimizers} use LLMs to directly generate solutions to optimization problems without calling traditional solvers through prompt optimization to improve performance.
The approach is limited to small problems since 
the performance of LLMs degrades as the input context grows, even for explicitly long-context models \citep{liu2023lostinthemiddle}. 

\paragraph{Chatbots for Optimization.} In a recent paper, \citet{chen2023diagnosing} developed a chatbot to help users detect and fix infeasible optimization problems expressed in \texttt{Pyomo} code and servers as an AI assistant rather than as a solver. \citet{li2023large} designed a chatbot to answer natural-language queries about an optimization model. \citet{mindoptCopilot} also developed a chatbot to facilitate optimization modeling, but there is no public paper or documentation available on it.

\paragraph{Benchmark-driven Optimization Modeling.} 
More closely related to our approach, \cite{ramamonjison2023nl4opt} introduced a dataset of 1101 natural language representations of LP problems. They proposed a two-stage mapping from the natural-language representation to the problem formulation using an intermediate representation. \cite{ramamonjison-etal-2022-augmenting} designed a system to simplify and improve the modeling
experience for operations research, but did not offer an end-to-end solution. \cite{anonymous2024chainofexperts} presented a multi-agent cooperative framework to automatically model and program complex operation research (OR) problems, and evaluated it on NL4Opt and another more complex dataset they curate. In terms of traditional MILP benchmarking, it should be noted that \texttt{MIPLIB} is widely recognized as a benchmark for evaluating the performance of MILP solvers. \texttt{MIPLIB} offers a diverse collection of MILP instance realizations that are, for the most part, detached from their original formulations. This paper focuses primarily on the modeling aspects of MILPs and therefore does not have a direct correlation with \texttt{MIPLIB}.

\section{Methodology}
\label{methodology}
\begin{figure*}[t!]
    \centering
    \includegraphics[width=\textwidth]{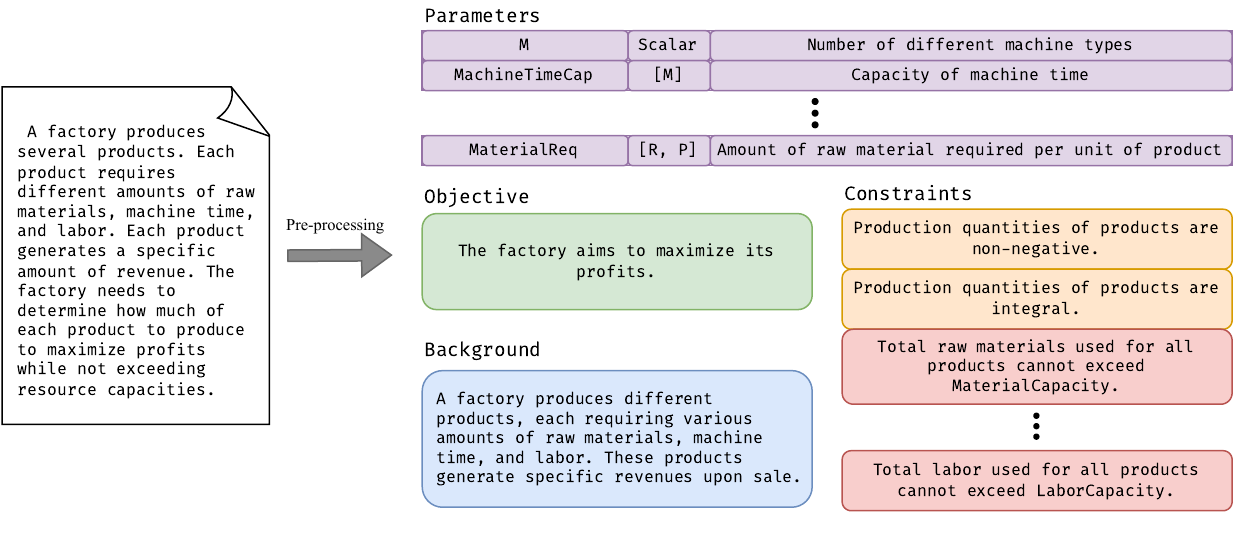}
    \caption{OptiMUS preprocesses natural language representations of a problem into a modular state. The components of the modular state are: \textbf{1)} parameters and their shape, \textbf{2)} objective, \textbf{3)} background and context, and \textbf{4)} implicit and explicit constraints.}
    \label{fig:preprocess}
\end{figure*}

\begin{figure*}
    \centering
    \includegraphics[width=\textwidth]{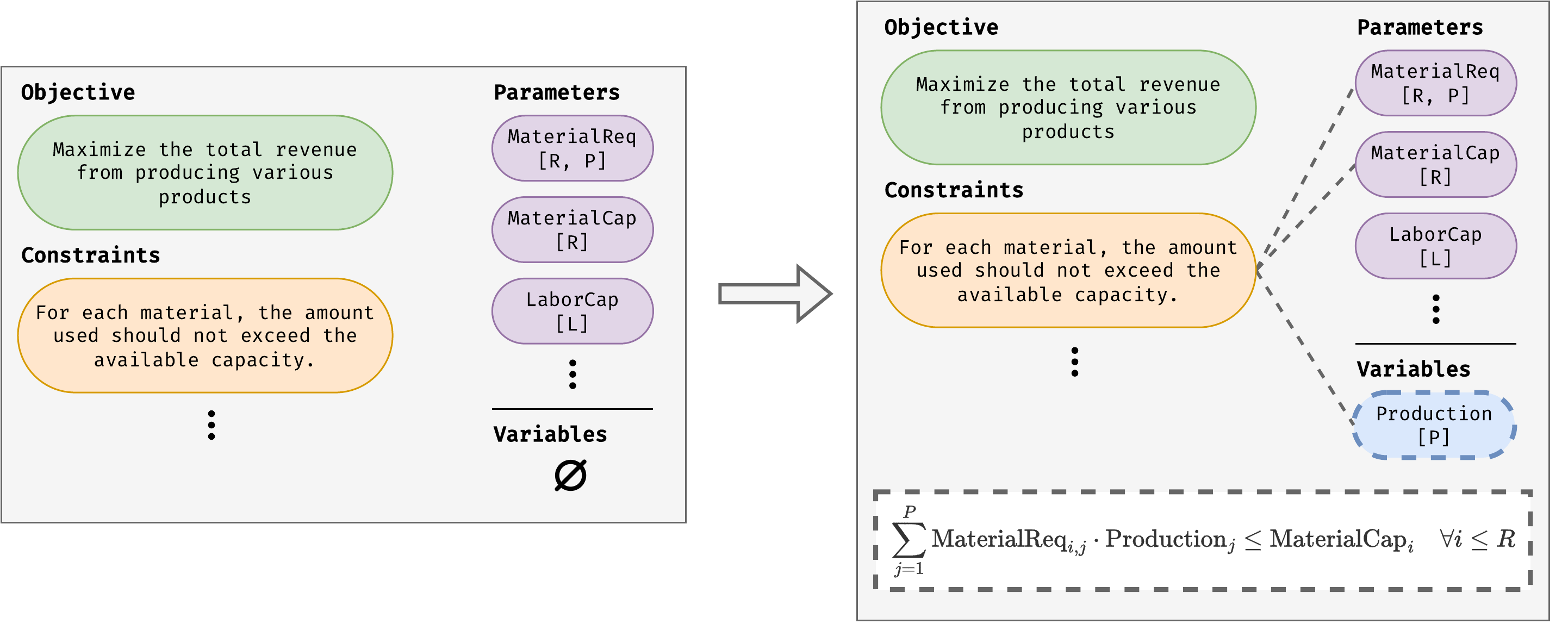}
    \caption{The formulation process for a single constraint. The formulation agent identifies any parameters and variables appearing in the constraint, including new variables that it may need to define. It defines new variables as needed, updates the connection graph which records which constraints use which parameters and which variables, and annotates the constraint with a \LaTeX~ formulation. (dashed lines represent new connections and variables) 
    }
    \label{fig:formulation-step}

\end{figure*}

This section details the design of OptiMUS. See \cref{fig:agent-team} for an illustration. 
The problem presented in Figure \ref{fig:preprocess} serves as a running example. 
OptiMUS starts with a natural language description of the optimization problem. The problem is first preprocessed to extract the parameters, constraints, objective function, and background information. 
Then OptiMUS uses a multi-agent framework to process and solve the structured problem.
\cref{prompts} includes all prompts used in OptiMUS.
For brevity, we use the word \emph{clause} to refer to a constraint or objective.

\subsection{Structured Problem}

The OptiMUS preprocessor converts a natural language description of the problem into 
a \textit{structured problem} (\cref{fig:preprocess}) with the following components:

\begin{itemize}[leftmargin=8pt]
    \item \textbf{Parameters}: A list of parameters of the optimization problem. Each parameter has three components: \textbf{1)} symbol, \textbf{2)} shape, and \textbf{3)} text definition.
    OptiMUS can choose symbols, infer the shape, and define the parameters if they are not explicitly included in the problem statement. 
    Importantly, numerical data that may be included in the problem statement is omitted from the parameters and stored for later use. This ensures that the parameters are short and easy to include in future prompts.  

\item \textbf{Clauses}: A list of the \emph{clauses} (objective and constraints) of the optimization problem. The preprocessor initializes each clause with its natural language description. Later these clauses will be augmented with {\LaTeX} formulations and code as well.
 
    \item \textbf{Background}: A short string explaining the real-world context of the problem. This string is included in every prompt to improve common sense reasoning.

\end{itemize}

The preprocessing uses three prompts:
the first prompt extracts the parameters,
the second segments the problem into objective and constraints,
and the third eliminates redundant (e.g., two restatements of the constraint that production quantity is nonnegative), unnecessary (such as facts about the problem parameters, e.g., that price is nonnegative), 
and incorrect constraints (e.g., production quantity must exactly equal demand).
The second step can also be a challenge:
for example, in  the factory example shown in \cref{fig:preprocess}, the production amount for each product should be a positive value, but this is not stated explicitly.

\label{agent-team}



\begin{algorithm}[t]
   \caption{Workflow of OptiMUS}
   \label{alg:flow}
    \begin{algorithmic}[1]
        \State {\bfseries Input:} Natural language description of problem $\mathcal P$
        \State $P^{(0)} \leftarrow \textsc{Preprocess}(\mathcal P)$
        \State Initialize $\textup{msg} \leftarrow$  ``"
        \State Initialize $\textup{conversation} \leftarrow  []$
        \For{$t=1,\ldots$}
        \State $\textsc{agent},  \textup{task} \leftarrow \textsc{Manager}(\textup{conversation})$
        \State $P^{(t+1)}, \textup{msg} \leftarrow \textsc{agent}(P^{(t)}, \textup{task})$
        \State 
        $\textup{conversation} \pluseq \textup{msg}$
        \State{\textbf{if} $\textup{msg} = \textup{Done}$ \textbf{then} break}
        \EndFor\\
        \textbf{end}
    \end{algorithmic}
\end{algorithm}

\subsection{Agents}
After preprocessing, OptiMUS defines problem variables, formulates and codes each clause. To ensure consistency of the formulations, OptiMUS constructs and maintains a connection graph to record which variables and parameters appear in each constraint. This connection graph is key to performance and scalability of OptiMUS, as it allows the LLM to focus only on the relevant context for each prompt, generating more stable results.
The list of variables and the {\LaTeX~} formulations and code are initially empty;
when all clauses are formulated, programmed, and validated, the process is complete.

\paragraph{Manager.} Inspired by \cite{wu2023autogen}, OptiMUS uses a manager agent to coordinate the work of formulation, programming, and evaluation, recognizing that these steps may need to be repeated to ensure consistency and correctness (see \cref{alg:flow}). At each step, the manager looks at the conversation so far and chooses the next agent (formulator, programmer, or evaluator) to process the problem. 
The manager also generates and assigns a \textit{task} to the chosen agent, for example: 

\begin{center}
\textit{Review and fix the formulation of the objective.}    
\end{center}

\paragraph{Formulator.} \label{sec:formulator}
The formulator agent must:
\begin{enumerate}[leftmargin=15pt]
    \item Write and correct mathematical formulations for variables and clauses.
    \item Define new variables and auxiliary constraints.
    \item  Update the links in the connection graph.
\end{enumerate}

If the assigned task is to formulate new clauses, the formulator iterates over the clauses that have not yet been formulated and generates new formulations for them. During this process, it will also define auxiliary constraints and new variables when necessary. Moreover, it decides which parameters and variables are related to the clause (see \cref{fig:formulation-step}). This information is used to update the connection graph. On the other hand, if the task is to fix incorrect formulations reported by the evaluator or the programmer, the agent iterates through the clauses marked as incorrect, fixes their formulations, and updates the connection graph. OptiMUS also has an extra modeling layer that captures special model structures (e.g., special-ordered-set and indicator variables) and we leave a more detailed discussion to the Appendix \ref{sec:opttechnique}.

\paragraph{Programmer.}
The responsibility of the programmer agent is to write and debug the solver code. 
When the programmer is called by the manager, it first reads the task. If the task is to program new clauses, the agent iterates over the clauses that have not yet been coded and generates code from their formulations. If the task is to fix incorrect formulations reported by the evaluator, the agent iterates through the clauses marked as bogus and fixes their codes. 

In our experiments, the programmer uses Python as the programming language and \texttt{Gurobi} as the solver. OptiMUS can target other solvers and programming languages as long as they are supported by the LLM.


\paragraph{Evaluator.}
The evaluator agent's responsibility is to execute the generated code on the data and to identify any errors that occur during the execution.
If evaluator faces a runtime error, it flags the variable or clause with the bogus code and responds to the manager with appropriate explanation of the error. The information will later be used by the other agents to fix the formulation and debug the code.

\subsection{The connection graph}
\begin{figure*}
    \centering
    \includegraphics[width=\textwidth]{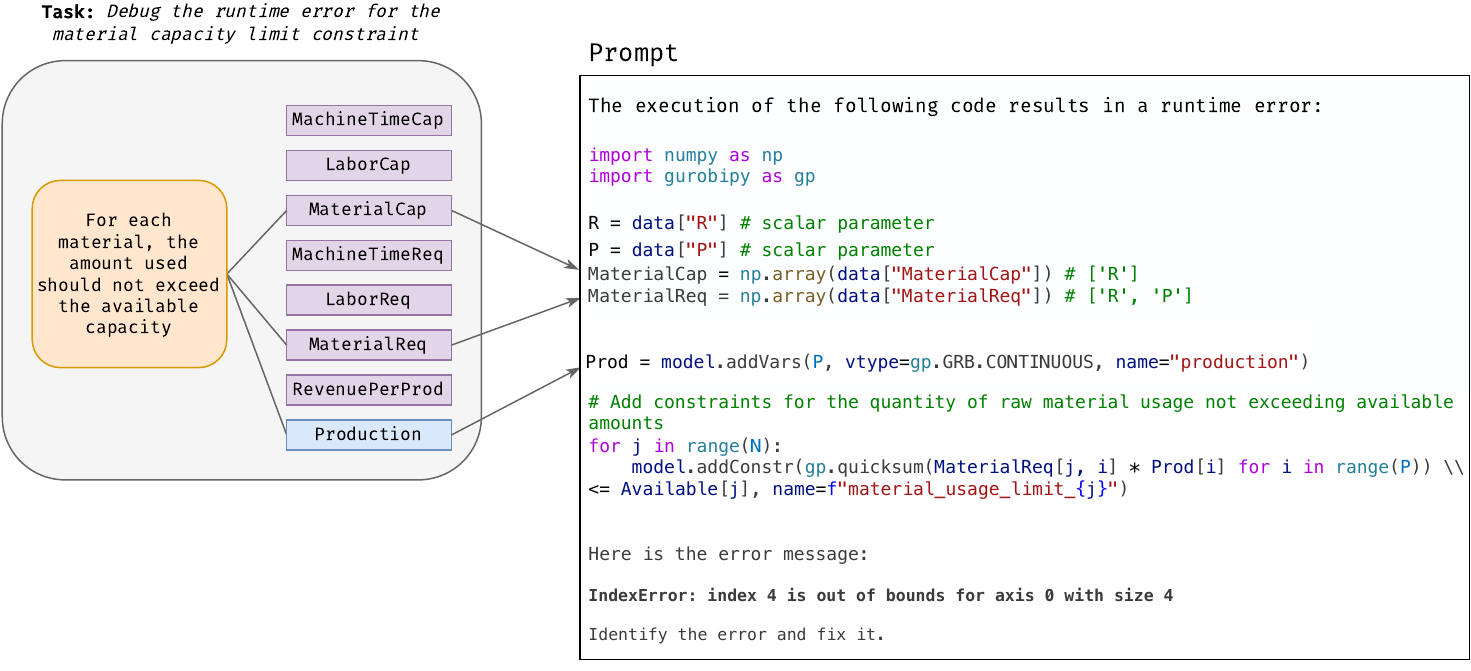}
    \caption{OptiMUS uses the connection graph to extract and use only the relevant context in each prompt. In this example, the programmer agent fetches the context via the connection graph to debug a bogus constraint code. Without the graph, the LLM would have needed to process the whole code, including the code for the other parameters, variables, constraints, and the objective.
    }
    \label{fig:graph-debug}

\end{figure*}

Recall from \cref{sec:formulator} that OptiMUS maintains a connection graph over constraints, objectives, parameters, and variables. 
OptiMUS uses this graph to retrieve the relevant context for each prompt so prompts remain short. 
This graph is used also to generate and debug code and to correct wrong formulations. 
\cref{fig:graph-debug} provides an example.

\section{Experiments}


In this section, we conduct a comprehensive evaluation of OptiMUS. We begin by detailing the datasets used in our experiments
and showcase the superior performance of OptiMUS across these datasets, highlighting its strengths. 
An ablation study demonstrates the impact of different system components on our results,
and a sensitivity analysis probes the internal dynamics of OptiMUS. 
We conclude this section by identifying failure cases and potential areas for further improvement.

\label{experiments}

\subsection{Dataset}
\label{section:dataset}

\textbf{NL4OPT}. This dataset is a collection of 1101 easy linear programming problems proposed as part of the NL4OPT competition \cite{ramamonjison2023nl4opt}. The dataset contains a natural language description of each problem, along with an annotated \emph{intermediate representation} that lists parameters, variables, and clauses.

\textbf{ComplexOR}. ComplexOR is a collection of 37 complex operations research problems in a variety of application domains \cite{anonymous2024chainofexperts}. At the time of writing this paper, the publicly available version of this dataset is incomplete. 
We gathered 21 problems from the ComplexOR dataset to use in our experiments by 
augmenting the problems that lack data with synthetic data. This modified dataset is available in our supplementary materials.

\textbf{NLP4LP}. As shown in \cref{table:datasets},
existing datasets for natural language optimization modeling lack problems with long descriptions. Real-world problems often are much longer, see e.g.~\cite{Challenge3}. 
To address this issue, we create NLP4LP (Natural Language Processing for Linear Programming), a benchmark consisting of \revised{54 LP and 13 MILP problems (67 instances in total)}. NLP4LP problems are drawn from textbooks and lecture notes on optimization \citep{intro_to_opt, model_building, lectures_in_lp_modeling}, including facility location, network flow, scheduling, portfolio management, and energy optimization problems. These resources were created before 2021, so it is possible parts of these books have been 
used to train LLMs. 
However, none of these textbooks includes code. Moreover, our results show that LLMs still find it challenging to formulate and solve these problems.
For each instance, NLP4LP includes the description, a sample parameter data file, and the optimal value, obtained either from the textbook solution manual or by solving the instance by hand.
Together, NLP4LP and ComplexOR offer a variety of challenging optimization problems with different lengths, facilitating the research on automated optimization modeling. 

\subsection{Overall Performance}
To evaluate the overall performance of OptiMUS, we compare it with standard prompting, Reflexion, and Chain-of-Experts (CoE) \cite{shinn2023reflexion, anonymous2024chainofexperts}. Reflexion is the highest-performing general-purpose framework and CoE is the state-of-the-art method for natural-language optimization modeling. Three main metrics have been used in the literature: accuracy, compilation error (CE) rate, and runtime error (RE) rate. However, a method can generate a totally irrelevant short code that runs, or fix runtime and complication errors by completely removing relevant sections of the code. Hence, we only compare the models' accuracy. Results are presented in \cref{table:performance}. OptiMUS outperforms all other methods in all datasets by a large margin. 
This remarkable performance improvement highlights the importance of modularity and structure compared to a single prompt to solve complex problems using LLMs.
The next experiments clarify which features of OptiMUS  contribute to its good performance.

\subsection{Ablation Study}

\cref{table:ablation} shows the impact of debugging and of the choice of LLM on the performance of OptiMUS. One interesting observation is the significant performance drop that occurs when smaller LLMs are used instead of GPT-4. The first reason is that the OptiMUS prompts are on average longer than the other methods and involve more complicated reasoning. Smaller LLMs are worse at reasoning \cite{wang2023far, openai2023gpt4}.  The second reason is the novel and modular structure of OptiMUS's prompts. Prompts used in the other methods mostly adhere to a questions answering format that is abundant in the public domain (e.g. posting the whole bogus code snippet and asking for the correct version is common on StackOverflow, or writing the whole problem description and then the complete formulation is common in optimization textbooks). However, in OptiMUS, the prompts are more complex and not common in human-human interactions. Smaller LLMs have limited generalization and reasoning abilities and, therefore, show poor performance on such prompts \cite{openai2023gpt4}.
Fine-tuning smaller models on these novel prompt templates might improve their performance and reduce the cost of 
running a system like OptiMUS. 

We also evaluated a version of OptiMUS which uses GPT-3.5 for the manager and GPT-4 for the other agents. We can see that in NL4OPT the difference in performance is small. The reason is that most instances of NL4OPT are solved with a simple chain of formulation-programming-evaluation. However, in ComplexOR and NLP4LP where more complicated interactions between agents are required, the manager's importance becomes more visible. Moreover, we did experiments in which the debugging feature of the programmer agent was disabled. Similarly to the manager, we see that debugging is more important in more complicated datasets.

\begin{table}[t]
\caption{Performance of OptiMUS and the baselines using GPT4.}
\label{table:performance}
\begin{center}
\begin{small}
\begin{tabular}{lccc}
\toprule
& NL4OPT & ComplexOR & NLP4LP \\
\midrule
Standard  & 47.3\% & 9.5\% & 35.8\% \\ 
Reflexion & 53\% & 19.1\% & 46.3\%\\
CoE & 64.2\% & 38.1\% & 53.1\% \\
\textbf{OptiMUS (Ours)} & \textbf{78.8\%} & \textbf{66.7\%} & \textbf{72.0\%} \\
\bottomrule
\end{tabular}
\end{small}
\end{center}
\end{table}

\begin{table}[t]
\caption{Ablation study of OptiMUS.}
\label{table:ablation}
\begin{center}
\begin{small}
\begin{tabular}{lccc}
\toprule
 & NL4OPT & ComplexOR & NLP4LP \\
\midrule
OptiMUS (GPT-4) & 78.7\% & 66.7\% & 71.6 \% \\
\midrule
w/o debugging & 72.3\% & 57.1\% & 58.2\% \\
w/ GPT-3.5 Mngr & 74.9\% & 52.4\% & 53.7\% \\
w/ GPT-3.5 & 28.6\% & 9.5\% & 14.4\% \\
w/ Mixtral-8x7B & 6.6\% & 0.0\% & 3.0\% \\

\bottomrule
\end{tabular}

\end{small}
\end{center}
\end{table}

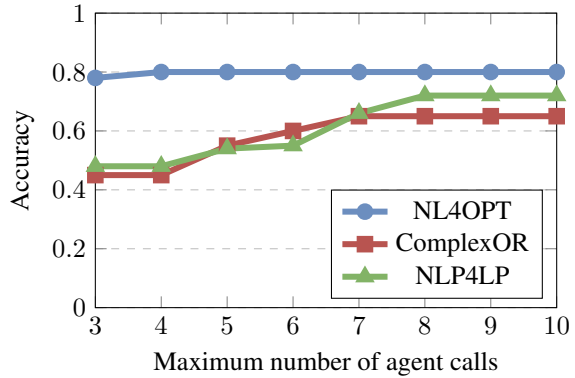
\begin{figure}[t]
    \centering
\begin{tikzpicture}
\begin{axis}[
    xlabel={Maximum number of agent calls},
    ylabel={Accuracy},
    xmin=3, xmax=10,
    ymin=0, ymax=1,
    xtick={3,4,5,6,7,8,9,10},
    ytick={0,0.2,0.4,0.6,0.8,1.0},
    legend pos=south east,
    ymajorgrids=true,
    grid style=dashed,
    width=0.45\textwidth,
    height=5.5cm
]

\addplot[color=NL4OPTcolor, mark=*,line width=0.8mm]
    coordinates {
    (3,0.78)(4,0.80)(5,0.80)(6,0.80)(7,0.80)(8,0.80)(9,0.80)(10,0.80)
    };
    \addlegendentry{NL4OPT}

\addplot[color=ComplexORcolor, mark=square*,line width=0.8mm]
    coordinates {
    (3,0.45)(4,0.45)(5,0.55)(6,0.60)(7,0.65)(8,0.65)(9,0.65)(10,0.65)
    };
    \addlegendentry{ComplexOR}

\addplot[color=NLP4LPcolor, mark=triangle*,line width=0.8mm]
    coordinates {
    (3,0.48)(4,0.48)(5,0.54)(6,0.55)(7,0.66)(8,0.72)(9,0.72)(10,0.72)
    };
    \addlegendentry{NLP4LP}
    
\end{axis}
\end{tikzpicture}
    \caption{OptiMUS can solve more problems on difficult datasets (ComplexOR, NLP4OPT) when more agent calls are allowed,
    demonstrating the importance of self-improvement.}
    \label{fig:steps}
\end{figure}

\subsection{Sensitivity Analysis}

\cref{fig:steps} shows how the maximum number of times the manager is allowed to select agents affects the accuracy. For NL4OPT, most problems are solved by selecting each of the formulator, programmer, and evaluator agents only once. However, for ComplexOR and NLP4LP, OptiMUS often makes mistakes at the beginning and iteratively fixes them by selecting the other agents multiple times. 

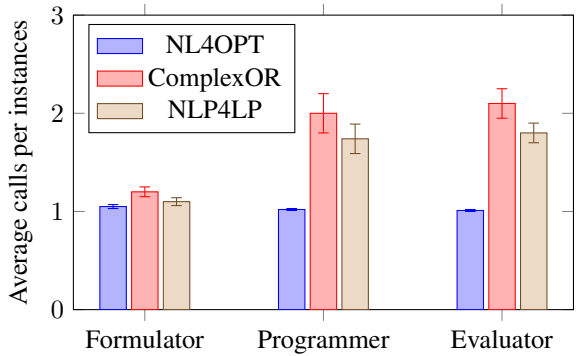
\begin{figure}
\begin{tikzpicture}
\begin{axis}[
    ybar, area legend,
    enlarge x limits=0.2, 
    legend style={
      anchor=north,
      legend image code/.code={ 
        \draw[##1,/tikz/.cd,yshift=-0.25em]
        (0cm,0cm) rectangle (3pt,0.8em);},
    },
    legend pos=north west,
    ylabel={Average calls per instances},
    symbolic x coords={Formulator, Programmer, Evaluator},
    xtick=data,
    ymin=0, ymax=3,
    height=5.5cm,
    width=0.48\textwidth,
    error bars/y dir=both,
    error bars/y explicit,
    ]

\addplot+[
    error bars/.cd,
    y dir=both,
    y explicit,
] coordinates {
    (Formulator,1.05) +- (0,0.02)
    (Programmer,1.02) +- (0,0.01)
    (Evaluator,1.01) +- (0,0.01)};
\addlegendentry{NL4OPT}

\addplot+[
    error bars/.cd,
    y dir=both,
    y explicit,
] coordinates {
    (Formulator,1.2) +- (0,0.05)
    (Programmer,2.0) +- (0,0.2)
    (Evaluator,2.1) +- (0,0.15)};
\addlegendentry{ComplexOR}

\addplot+[
    error bars/.cd,
    y dir=both,
    y explicit,
] coordinates {
    (Formulator, 1.1) +- (0,0.04)
    (Programmer, 1.74) +- (0,0.15)
    (Evaluator, 1.8) +- (0,0.1)};
\addlegendentry{NLP4LP}

\end{axis}
\end{tikzpicture}
\label{fig:agent_calls}
\caption{Average number of calls to each agent among solved problems. OptiMUS only requires one call per agent on the simple problems of NL4OPT. On the more complex datasets, it relies more heavily on the programmer to fix errors identified by the evaluator, but rarely improves by fixing formulation errors.}
\end{figure}

\cref{fig:agent_calls} shows the number of times each agent is selected per instance. As expected, the average selection frequency is higher in ComplexOR and NLP4LP. Moreover, programmer and evaluator agents are selected more often than the formulator. This bias is reasonable:

\begin{itemize}[leftmargin=10pt]
\item Coding errors are more common. LLMs often generate code with trivial bugs that are easy to fix. In OptiMUS, the programmer agent fixes such bugs. 

\item Coding errors are easier to identify and fix. In contrast, identifying bugs in the formulation require deeper reasoning and is harder. Hence the manager in OptiMUS is prompted to prioritize fixing the code before considering errors in the formulation. The formulator is only selected for debugging if the programmer claims that the code is correct. 
\end{itemize}

Hence in our experiments, we observe the programmer is selected more often than the formulator.

\cref{table:promptInfo} shows the average prompt length of OptiMUS and CoE for different data sets.  
Observe that the prompt length for OptiMUS barely changes across datasets, while the prompt length for CoE increases on more challenging datasets. The reason is the modular approach, which allows OptiMUS to extract and process only the relevant context for each LLM call. Unlike non-modular methods, OptiMUS can scale to larger and longer problems.

\subsection{Failure Cases}

To understand its strengths and weaknesses, we analyze the most common reasons why OptiMUS fails (\cref{table:falure-cases}). 
We categorize failure cases into the following groups:

\begin{itemize}[leftmargin=15pt]
    \item Missing or wrong constraints: OptiMUS generates wrong constraint in the preprocessing step (e.g., $\text{price} \geq 0$ where price is a parameter), or fails to extract all of the constraints from the description. 
    \item Incorrect model: OptiMUS tackles the problem with an incorrect mathematical model (e.g., defining binary variables for visiting cities instead of links in TSP).
    \item Coding error: OptiMUS does not generate error-free code even after debugging. 
    Coding errors often occur when the LLM is confused by the language used (e.g., in the ``prod'' problem in ComplexOR, the description explicitly refers to ``parameters'' and ``variables'').
\end{itemize}

We normalize the failure rates to sum to 1.0. Incorrect modeling is more common on datasets with more complicated problems, while on the easier dataset NLP4OPT, the model is less likely to be wrong.

Understanding and interpreting the problems is also challenging for LLMs, 
resulting in formulations with missing constraints and wrong constraints. 
Fine-tuning might improve the performance of LLMs on this task, and is an important direction for future research.


\section{Conclusion}
\label{conclusion}

How can LLMs collaborate and divide work in order to achieve complex goals?
This paper interrogates this question in the domain of optimization
and showcases the importance of modular structure. 
We develop OptiMUS, 
a modular LLM-based agent designed to formulate and solve optimization problems from natural language descriptions. 
Our research serves as a proof-of-concept, illustrating the potential for automating various stages of the optimization process by combining LLMs with traditional solvers.
To showcase the performance of OptiMUS, we released NLP4LP, a dataset of long and challenging optimization problems to demonstrate the efficacy of the techniques implemented within OptiMUS. 
OptiMUS achieves SOTA performance across all existing datasets, and scales to problems with large amounts of data and long descriptions. 

Real-world optimization problems are often complex and multifaceted. 
Developing LLM-based solutions for these problems requires domain-specific considerations, including integrating existing optimization techniques to leverage problem structure. We are at the early stages of this research, but anticipate significant developments that will enable these systems to address more complex, industrial-level problems. It is interesting to notice that the challenge of using AI for an applied domain is much larger in safety-critical domains such as self-driving, which demand extremely high accuracy, than in domains where AI can function as an assistant and where answers are easy to check, as in theorem-proving or optimization. Here, AI systems with moderate accuracy can still usefully augment human effort.


\paragraph{Future directions.} 
Smaller LLMs are faster and cheaper, but our experiments indicate that they perform poorly in optimization modeling out-of-the-box. Identifying which prompts might benefit from fine-tuned small models and which require large (and expensive) LLM calls is an important topic for future research. 
Furthermore, we believe that integrating user feedback into the process can improve the performance of agents on natural-language optimization modeling. Studying interactions between such agents and their users is an exciting avenue. 
Another important direction is to automatically select the best solver based on a comprehensive evaluation of both accuracy and runtime requirements.
Additionally, it would be interesting to see how the modular LLM structure presented here can be enhanced using reinforcement learning to teach the manager how to choose the next agent.


\begin{table}
\caption{CoE requires longer prompts on difficult datasets, while OptiMUS barely increases its prompt length.\label{table:promptInfo}}
\begin{center}
\begin{small}
\begin{tabular}{lccc}
\toprule
 & NL4OPT & ComplexOR & NLP4LP \\
\midrule
\midrule
CoE &2003 $\pm$ 456 & 3288 $\pm$ 780 & 3825 $\pm$ 1002 \\
OptiMUS & 2838 $\pm$ 822 & 3241 $\pm$ 1194 & 3146 $\pm$ 1145\\
\bottomrule
\end{tabular}
\end{small}
\end{center}
\end{table}

\begin{table}
\caption{When OptiMUS fails, why?}
\label{table:falure-cases}
\begin{center}
\begin{small}
\begin{tabular}{lccc}
\toprule
Mistake & NL4OPT & ComplexOR & NLP4LP \\
\midrule
\midrule
Incorrect modeling & 43.0\% & 62.5\% & 53.8\%\\
Missing constraints & 36.0\% & 12.6\% & 15.4\%\\
Coding errors & 21.0\% & 24.9\% & 30.8\%\\
\bottomrule
\end{tabular}
\end{small}
\end{center}
\end{table}

\section{Impact Statement}
This paper presents work whose goal is to advance the field of optimization modeling. There are many potential
societal consequences of our work, none which we feel must
be specifically highlighted here.

\bibliography{paper}
\bibliographystyle{icml2024}


\newpage
\appendix
\onecolumn

\newcommand{\coloredText}[1]{\textcolor{blue}{#1}}



\section{Optimization Techniques}
\label{sec:opttechnique}
Optimization solvers exploit problem-specific structure to improve performance when solving MILPs \cite{gamrath2016structure}
and often provide a customized interface for these special structures. 
Using the interface not only reduces the complexity of (and potential for errors in) auxiliary variables or constraints, but also informs the solver about the existence of structure that can be exploited to solve the problem faster. 
Moreover, the solver's performance can suffer when these structures are not signaled in the model. For example, a bad choice of big-M coefficient when reformulating an indicator variable can reduce the strength of the linear relaxation. Typical examples of structure include Special Ordered Set (SOS) \cite{beale1976global}, indicator variables, and general constraints \cite{bertsimas1997introduction}.

Although state-of-the-art optimization solvers can detect some problem structures automatically, it works better to specify structure during problem formulation.
Hence the formulator is prompted to leverage advanced optimization techniques and structures, including \textbf{1)} Special Ordered Set. \textbf{2)} Indicator variable. \textbf{3)} General constraints. \textbf{4)} SAT and constraint programming problem. \textbf{5)} Totally unimodular problem detection.

\begin{figure}[t]
 \centering
 \includegraphics[width=\textwidth]{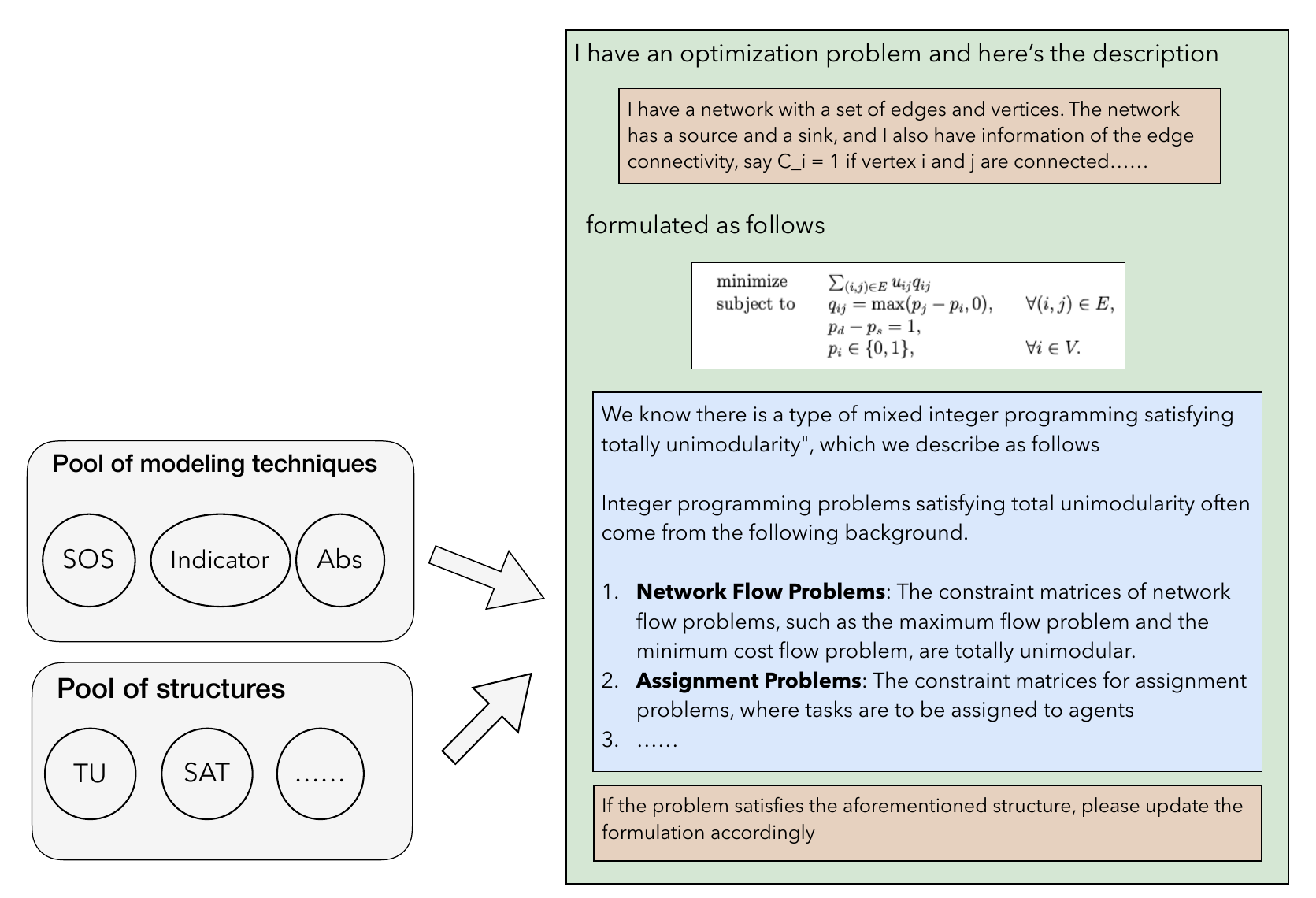}
 \caption{OptiMUS iterates through a pool of advanced optimization techniques \label{fig:opt-technique}}
 \end{figure}

OptiMUS  iterates through a sequence of ``cheatsheet'' prompts (Figure \ref{fig:opt-technique}), each corresponding to one of these structures. Within each prompt, the LLM is provided with the description of the structure, explained by an example illustrating how the structure should be exploited. The LLM is asked to decide whether the structure can be applied to the existing formulation. Upon identifying the appropriate structure, the formulation is adjusted to utilize the customized solver interface when available.

\section{prompts}
\label{prompts}

\subsection{Manager Prompt}
\includegraphics[width=0.95\textwidth]{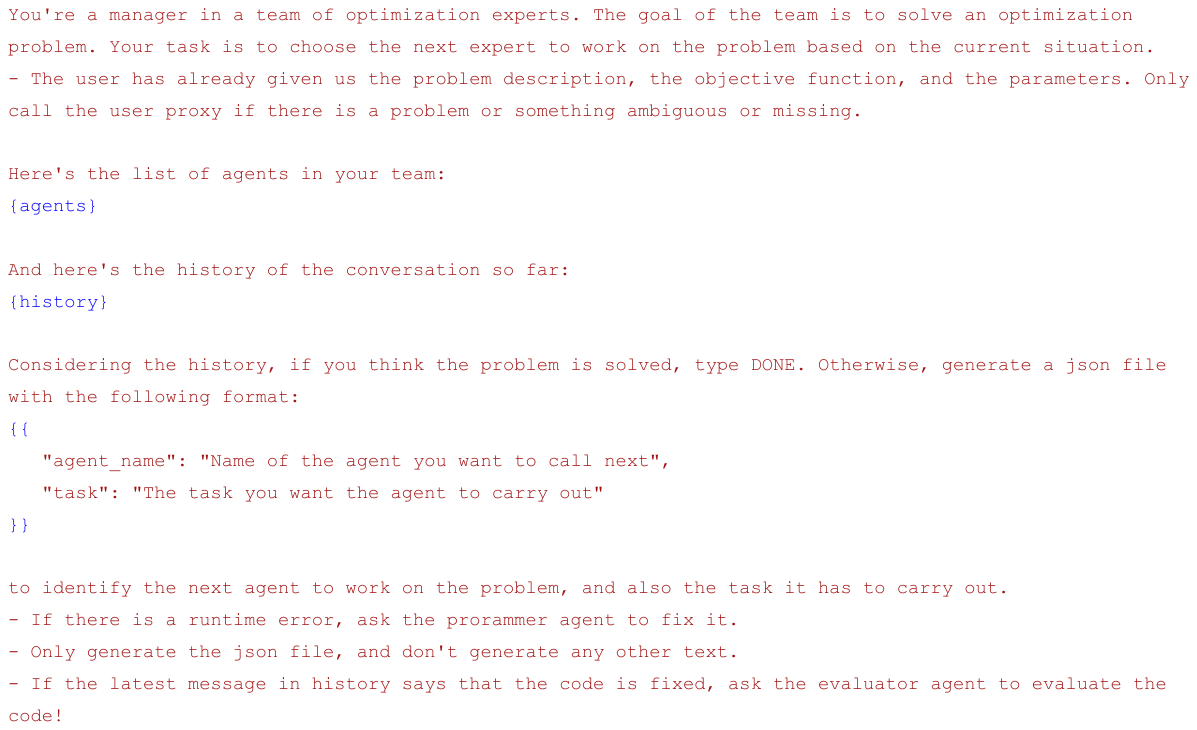}

\subsection{Formulation generation prompt}
\includegraphics[width=0.97\textwidth]{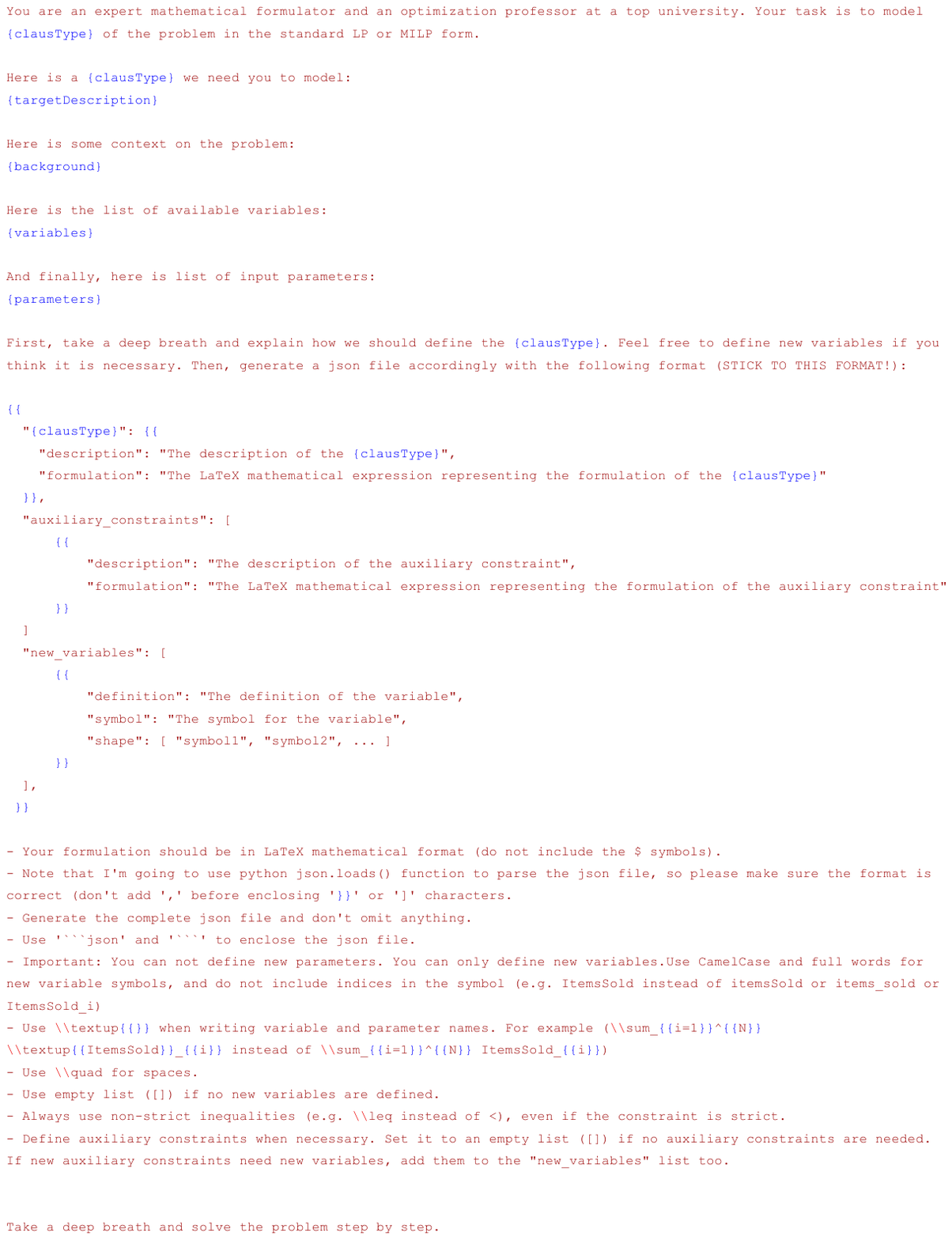}

\subsection{Formulation fixing prompt}
\includegraphics[width=0.97\textwidth]{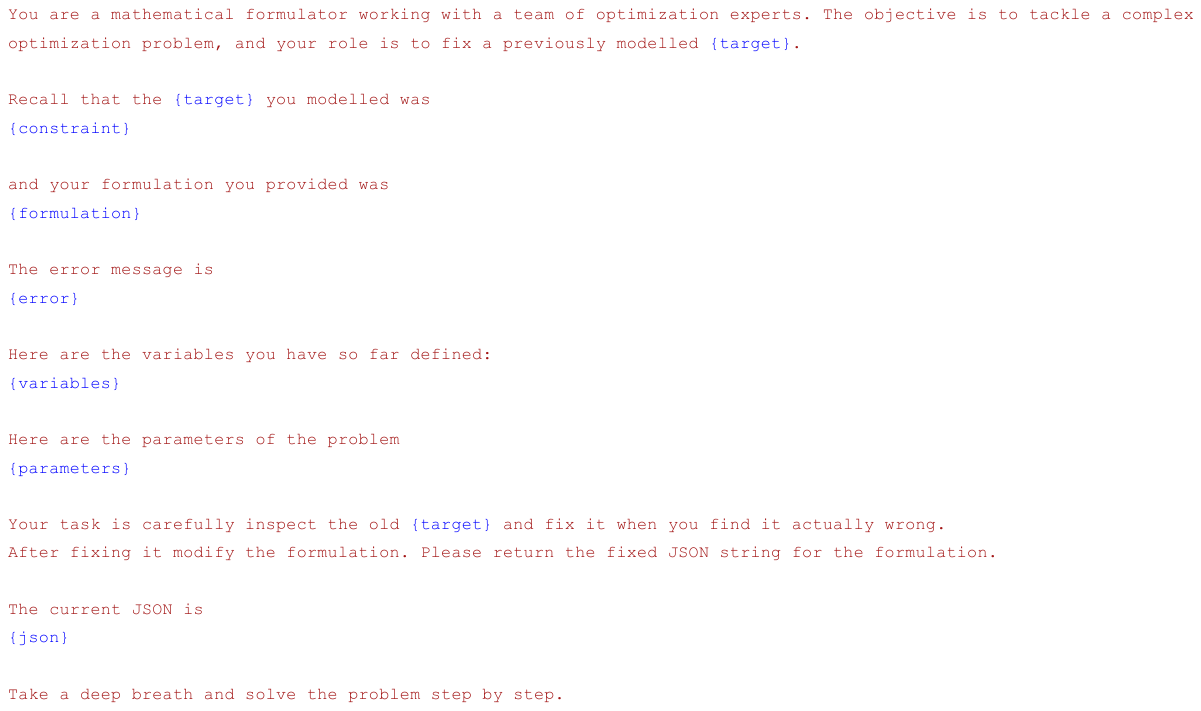}

\subsection{Clause Coding prompt}
\includegraphics[width=0.97\textwidth]{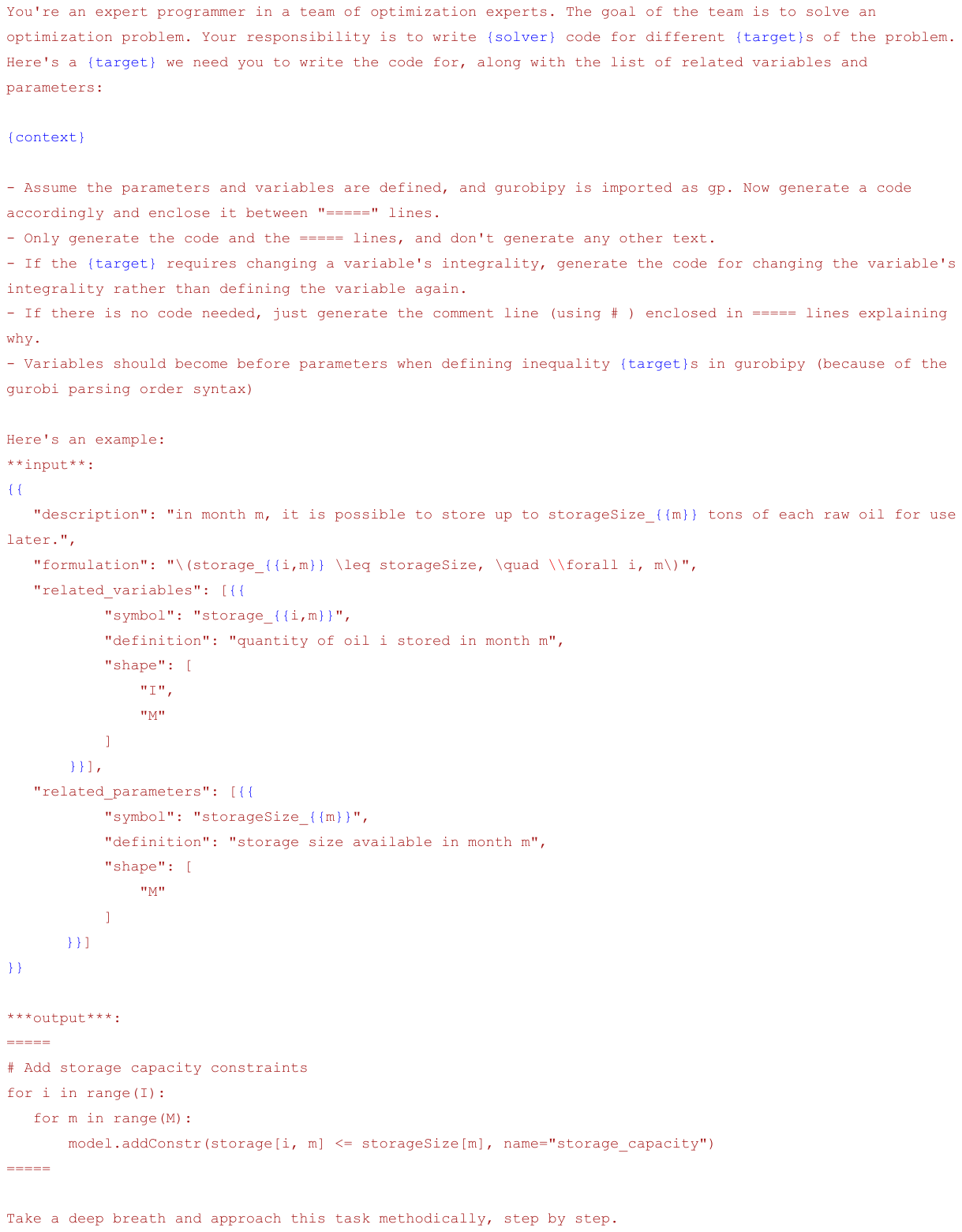}

\subsection{Variable coding prompt}
\includegraphics[width=0.97\textwidth]{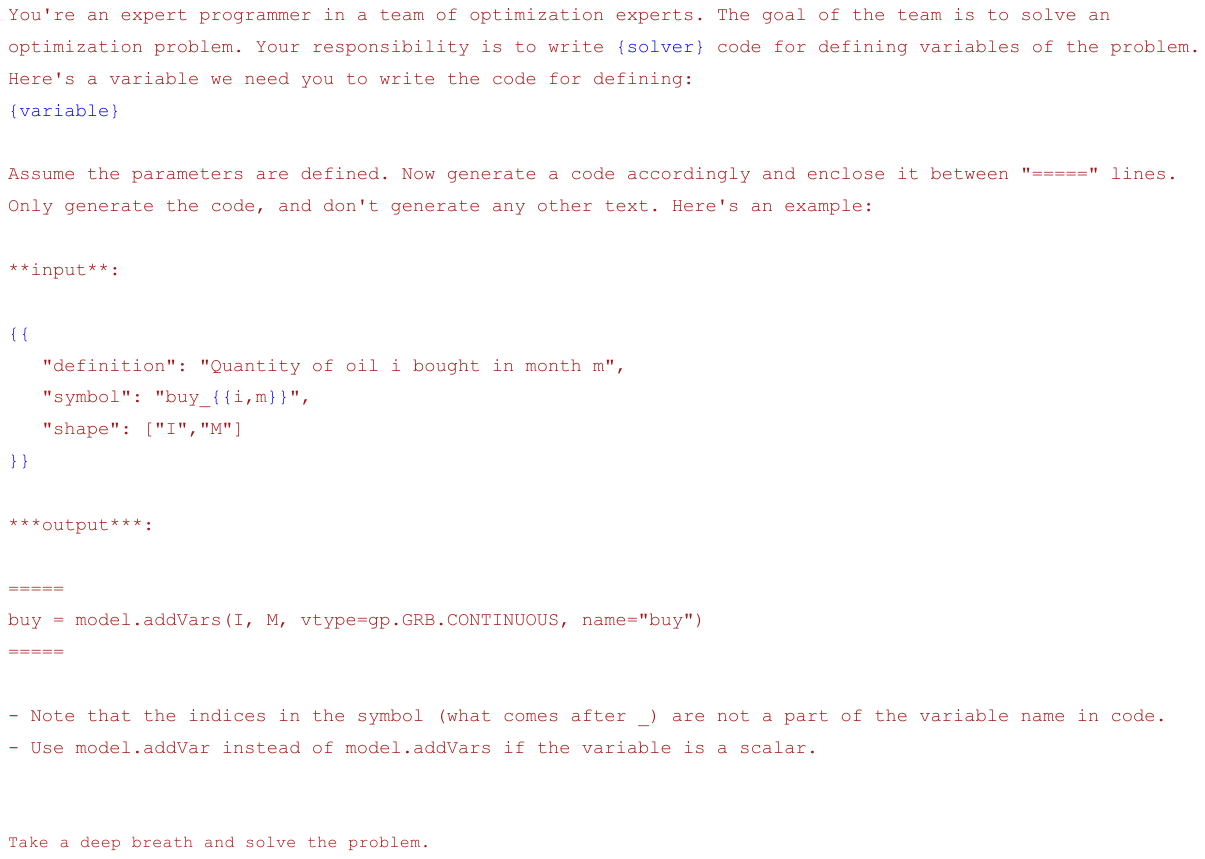}

\subsection{Debugging prompt}
\includegraphics[width=0.97\textwidth]{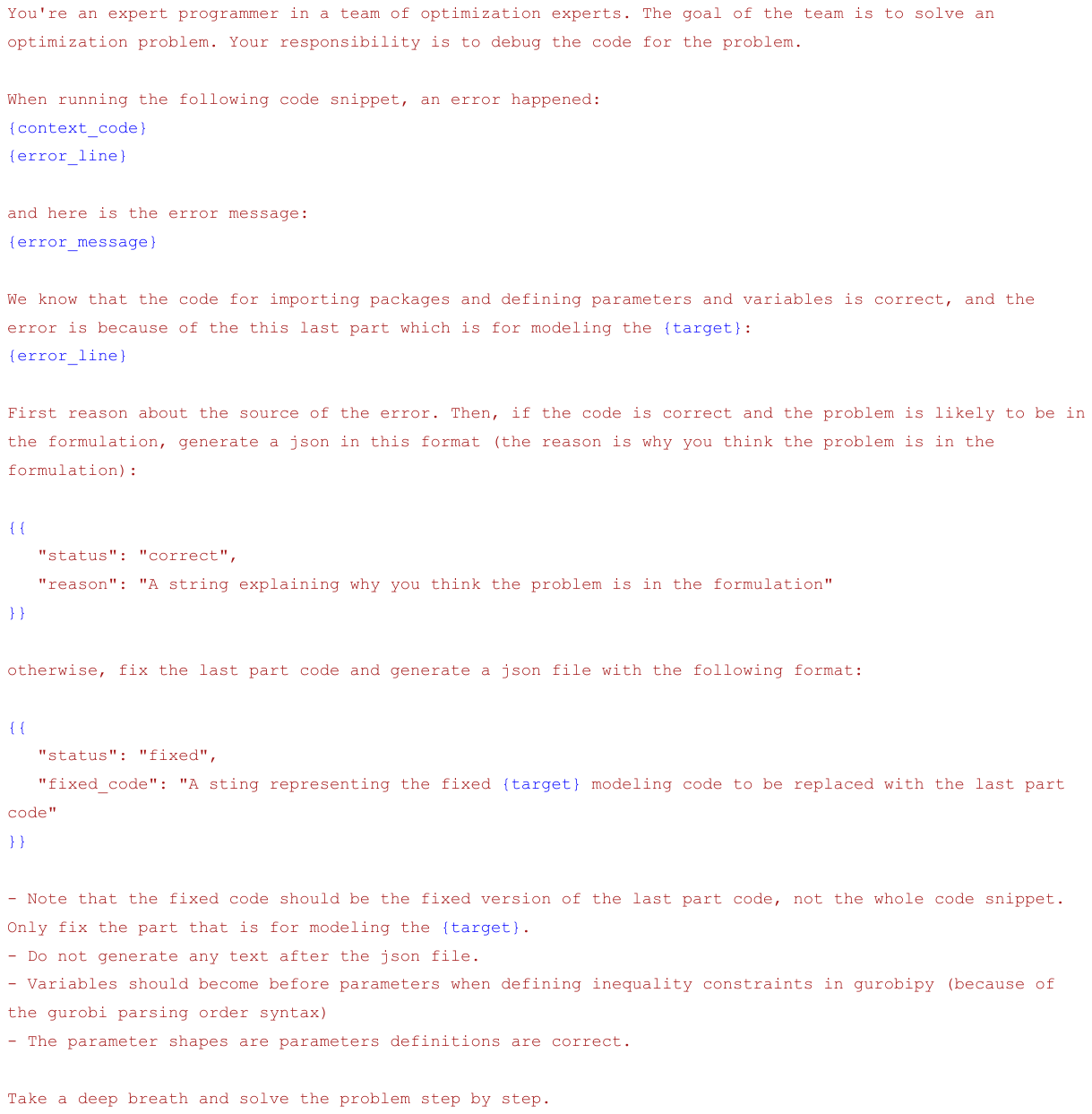}


\end{document}